\documentclass[journal]{IEEEtran}

\usepackage[dvipsnames]{xcolor}
\usepackage{graphicx}
\usepackage{subfig}
\usepackage{amsmath}
\usepackage{amsfonts}
\usepackage{lipsum}
\usepackage{tabularx}
\usepackage{url}
\usepackage{gensymb}

\usepackage[numbers]{natbib}

\begin{document}
\title{iFANnpp: Nuclear Power Plant Digital Twin for Robots and Autonomous Intelligence}

\author{Youndo~Do$^{1}$,
        Marc~Zebrowitz$^{1}$,
        Jackson~Stahl$^{1}$,
        and~Fan~Zhang$^{1,*}$
\thanks{$^{1}$ George W. Woodruff School of Mechanical Engineering, Georgia Institute of Technology}%
\thanks{$^{*}$ Corresponding author}%
}

\maketitle

\begin{abstract}
Robotics has gained attention in the nuclear industry due to its precision and ability to automate tasks. However, there is a critical need for advanced simulation and control methods to predict robot behavior and optimize plant performance, motivating the use of digital twins. Most existing digital twins do not offer a total design of a nuclear power plant. Moreover, they are designed for specific algorithms or tasks, making them unsuitable for broader research applications. In response, this work proposes a comprehensive nuclear power plant digital twin designed to improve real-time monitoring, operational efficiency, and predictive maintenance. A full nuclear power plant is modeled in Unreal Engine 5 and integrated with a high-fidelity Generic Pressurized Water Reactor Simulator to create a realistic model of a nuclear power plant and a real-time updated virtual environment. The virtual environment provides various features for researchers to easily test custom robot algorithms and frameworks.
\end{abstract}

\begin{IEEEkeywords}
Digital Twin, Robotics, Nuclear Power Plant, Artificial Intelligence 
\end{IEEEkeywords}

\IEEEpeerreviewmaketitle

\section{Introduction}

\IEEEPARstart{R}{obots} have emerged as potential transformative tools across many industrial control systems including nuclear power plants, driven by their incredible precision \cite{von2021precise, katsamenis2022simultaneous, pretto2020building}, enhanced productivity \cite{wu2022human, lagomarsino2022robot, johnson2022multi}, diversity in controls \cite{truong2021backstepping, he2020admittance, wang2022control}, and range of locomotion methods \cite{li2023design, taheri2023study, li2023aerial}. 

Robots have emerged as potential transformative tools across many industrial systems including nuclear power plants, driven by their incredible precision \cite{von2021precise, katsamenis2022simultaneous, pretto2020building}, enhanced productivity \cite{wu2022human, lagomarsino2022robot, johnson2022multi}, diversity in controls \cite{truong2021backstepping, he2020admittance, wang2022control}, and range of locomotion methods \cite{li2023design, taheri2023study, li2023aerial}. However, errors in robot control software can result in severe hardware damage or even human injury. In addition, real-world robot development and testing can be slow and expensive. To help mitigate these issues, digital twin platforms can be relied on for safe, rapid, and cost-efficient prototyping.
    
A digital twin is a solution to develop a realistic, virtual model of a system with continuous updates of its real-time data and behavior. It enables real-time monitoring and physics simulation, allowing operators to predict system behaviors. The increase in predictability improves nuclear power plant reliability by reducing plant failures. Additionally, a digital twin provides a platform for testing research ideas without accessibility constraints. Access to nuclear power plants is highly restricted due to their potential hazards and confidentiality. A digital twin is perfect for this environment since it presents the simulation of the nuclear power plants and produces data that can be further used for multiple functions. Additional follow-up research can be conducted, including simulation of various operations in nuclear power plants, optimization of the operations, prediction of outcomes using robots, and real-time monitoring of nuclear power plant component behaviors.

\begin{figure*}
    \centering
    \includegraphics[width=\textwidth]{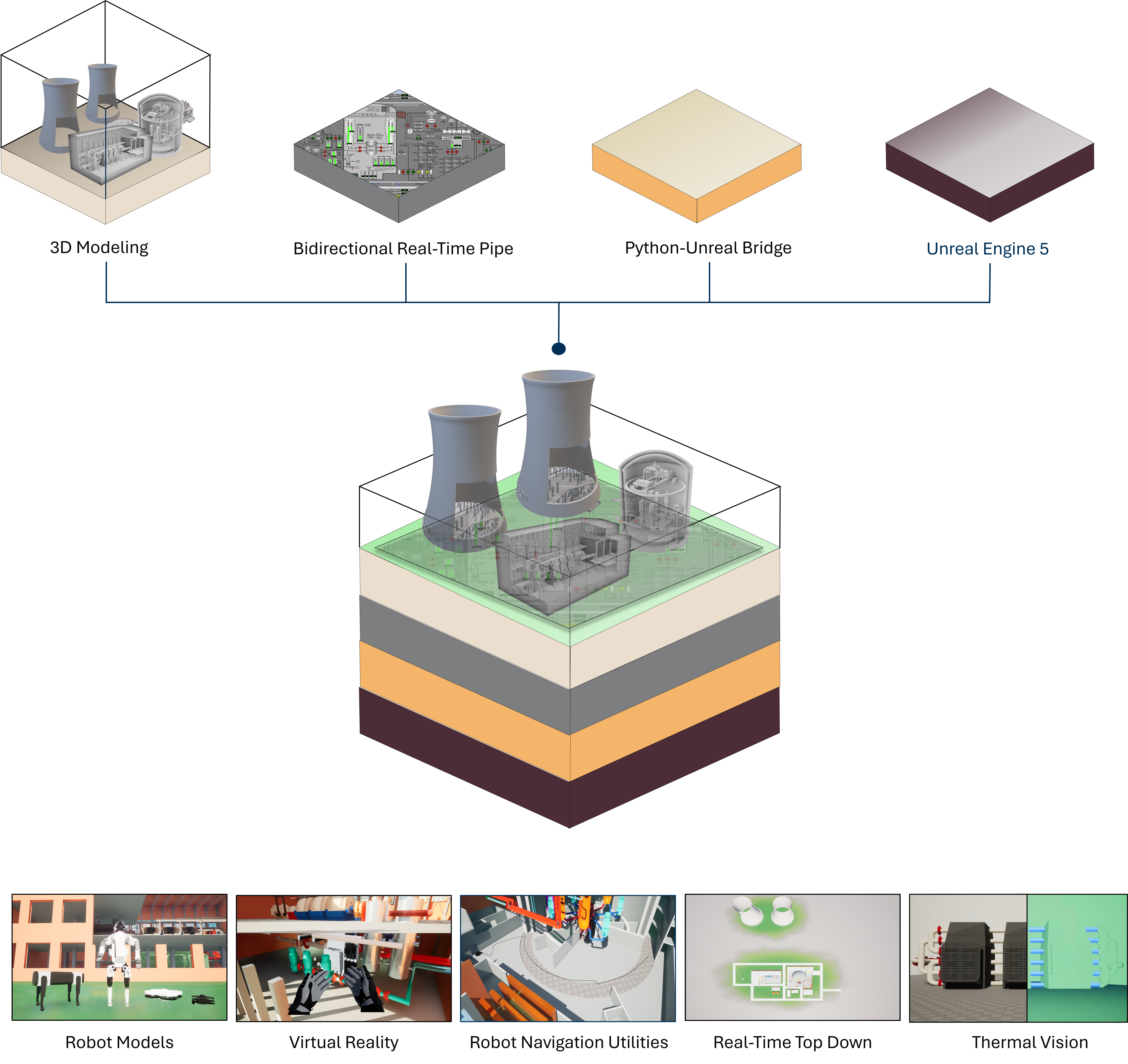}
  \caption{iFANnpp architecture and features}
  \label{fig:ifannpp-Architecture}
\end{figure*}

This paper introduces a nuclear power plant digital twin for robots and the development of artificial intelligence (AI), called iFANnpp. The digital twin is capable of supporting the development of advanced technologies such as machine learning (ML), reinforcement learning (RL) and autonomous robot controls. Figure \ref{fig:ifannpp-Architecture} illustrates the overview framework of iFANnpp, which includes the core digital twin features such as a bidirectional communication pipeline using a high-fidelity physics simulator for accurate nuclear power plant simulation and a Python-Unreal bridge for a user-friendly ML/RL testbed. This platform uses Unreal Engine 5 (UE5), a powerful 3D modeling and simulation software. It offers several key features:

\begin{enumerate}
        \item Python-Unreal Bridge: A communication pipeline between experimental research code (such as a RL control algorithm) and the digital twin environment. The UnrealCV plugin \cite{qiu2017unrealcv} is used to assist this pipeline. This feature allows integration of experimental Python code and an ecosystem of external Python libraries to interact with and receive feedback from the digital twin environment.
        \item Real-Time Mirroring: A bidirectional communication bridge between a third-party nuclear power plant physics simulator and the digital twin. This feature provides real-time updates to/from the digital twin and simulates realistic nuclear power plant responses based on the interaction from the proposed digital twin.
        \item Robot Models: Various 3D models for different tasks in the digital twin, such as wheeled, bipedal, quadruped, and aerial robots. This provides various models that practitioners can use and extend for their research.
        \item Virtual Reality (VR): A popular tool to simulate one's presence in virtual space. This enables the deployment of the proposed digital twin in a virtual reality environment, enabling immersive and smooth control of the interactive environment.
        \item Robot Navigation Utilities: A technique for tracing the movement of robots. Utilizing algorithms provided by researchers, the feature provides functionality for visualizing the path of a robot within the digital twin to view trajectories from a starting location to a target location.
        \item Real-Time Top-Down Perspective: Bird-eye view that provides a different perspective from the first- or third-person perspective. This feature captures and saves 2D top-view images of the system, providing a visual representation of its layout and structure, synchronized with the digital twin's time frame.
        \item Thermal Vision: Camera-based vision that displays a visual representation of a heat map of the surrounding environment. The feature enables thorough thermal analysis and monitoring for safety and performance optimization.
\end{enumerate}

The remainder of the paper is structured as follows: Section II introduces related research on industrial digital twins and associated robots. Section III discusses the modeling of iFANnpp. Section IV describes the various features presented in iFANnpp. Section V demonstrates potential use cases of the digital twin. Section VI presents an analysis on the performance of the digital twin. Section VII concludes the paper.

\section{Related Works}

\subsection{Industrial Digital Twin}

Digital twins have emerged as a pivotal technology in industrial systems, offering enhanced reliability and accuracy for simulating and testing real-world scenarios. They also provide a cost-effective and safe solution for training and remote operation. To deliver these benefits, it is crucial to accurately replicate the real-world environment with precise 3D modeling and seamless integration of real-time data.

Several studies have explored integrating advanced 3D models with real-time data to create accurate and functional digital twins. Sørensen et al. \cite{sorensen2022potentials} developed a digital twin using Unreal Engine 5 specifically for wind power analysis, allowing real-time simulation and turbine performance optimization. Garg et al. \cite{garg2021digital} implemented a virtual reality (VR) feature with Unity 3D to connect a virtual world to a physical FANUC robot, providing remote control and monitoring its operation. Erdei et al. \cite{erdei2022design} used Unreal Engine 4 to create a cyber-laboratory space for KUKA KR5 work area for training purposes. Metzner et al. \cite{metzner2020system} created a VR digital twin of a robot in an industrial control system, specifically for simulating programmable logic controllers (PLCs) in Unity. Gonzalez et al. \cite{gonzalez2024approach} modeled a laboratory room virtually and augmented it with reality capabilities using a mobile device, enhancing virtual interaction more realistically. Vairagade et al. \cite{vairagade2024nuclear} developed an NVIDIA Omniverse-based digital twin for nuclear power plants using the same Generic Pressurized Water Reactor (GPWR) simulator, a prior work that was limited to realistically modeling a single building within the nuclear power plant system. This work models all primary functional components of a pressurized water reactor (PWR) and offers a broader range of features, including bidirectional communication with the underlying simulator and the Python-Unreal bridge.

As listed above, many digital twin studies focus on modeling specific components or confined spaces within an industrial environment. The iFANnpp platform aims to expand the scope of digital twin technology by modeling an entire nuclear power plant to include a broader range of operational contexts. This approach has significant potential to deliver deeper insights into system performance and operational optimization, which are difficult to detect with individual components or processes. 

\subsection{Robots in Industrial Environments}
Multiple works have studied the use of robotics for industrial applications. Hobert et al. \cite{hoebert2023knowledge} developed a framework for executing robotic tasks to program industrial robots effectively. Ipiales et al. \cite{ipiales2023virtual} modeled an industrial Scara SR-800 robot in simulation, modeling the dynamics and kinematics. Bilancia et al. \cite{bilancia2023overview} described multiple robot controller configurations that aim to improve over traditional industrial robot configurations.  

While the former works focus on the modeling and control of stationary robots to enable fine-grained manipulation and automation for industrial applications, additional works feature mobile robotics. Mobile robotics offer the flexibility of movement, but can also add design complexity regarding power management and algorithmic control \cite{farooq2023power}. Naveen et al. \cite{naveen2023wireless} designed and implemented a solar-powered mobile robot with pick and place capabilities as well as control software for potential industrial automation tasks. Ghodsian et al. \cite{ghodsian2023mobile} provides a survey of industrial mobile robotics, highlighting the importance of their development for Industry 4.0 designs, their flexibility in terms of movement and coordination, as well as the complexities and challenges in their design to ensure valid function and safety. Buerkle et al. \cite{buerkle2023towards} discuses a potential industrial robot system architecture using mobile robots, called Industrial Robots as a Service, highlighting potential methods for programming mobile robots such as through explicit task programming or via human-robot interaction. Additionally, the work emphasizes the need for robot adaptability to environmental change and the critical importance of safety, primarily when humans work near autonomous machines.

iFANnpp provides a platform for simulating stationary and mobile robots specifically for nuclear power plant applications. The transition to deploying mobile robots for industrial applications presents many challenges, including the need for more complex algorithm development and a safe environment. Using iFANnpp, researchers can prototype mobile robot planning and control methods, confirming design and execution validity before deploying them to the real world.

\begin{figure*}
    \centering
    \includegraphics[width=7in]{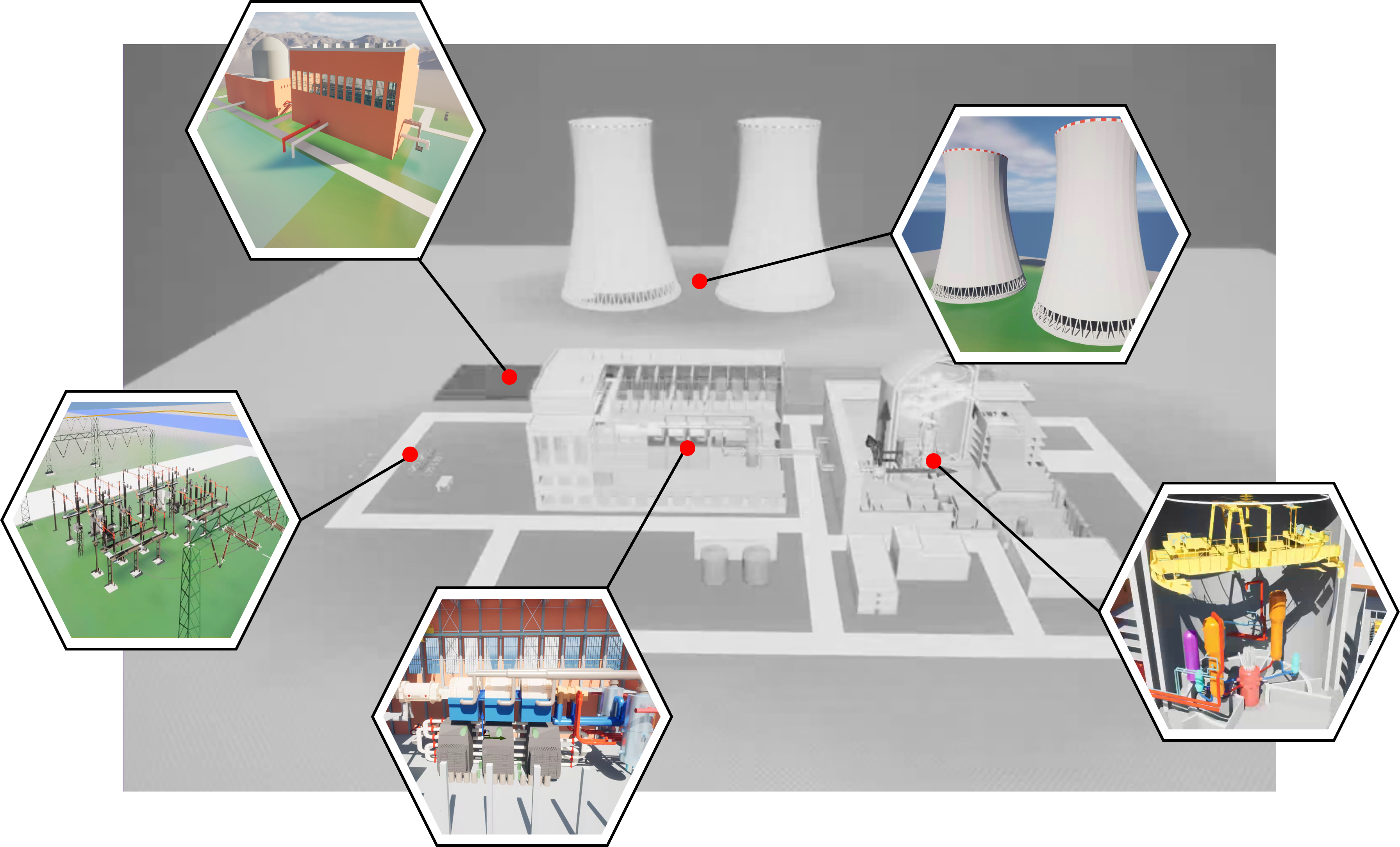}
    \caption{The overview of iFANnpp: Power output area, Water sources, Turbine hall, cooling towers, and reactor building are shown from left to right}
    \label{fig:iFANnpp-Overview}
\end{figure*}

\section {Digital Twin Modeling}

iFANnpp includes 3D models for the entire nuclear power plant, shown in Figure \ref{fig:iFANnpp-Overview}. A nuclear power plant is a complex industrial system, which makes it an ideal candidate for a comprehensive simulation platform. Nuclear power plants place a strong emphasis on safety and reliability constraints, which introduce many scenarios that are also relevant to other industrial control systems. This work utilized  basic models for an PWR from Encyclopedia \cite{EnergyEncyclopedia} and modified it and expanded it to synchronize a high fidelity full scope GPWR Simulator \cite{gpwr} developed by Western Services Corporation (WSC). The iFANnpp digital twin model is divided into four main zones: the reactor building, the turbine hall building, the water source and cooling towers, and the power area.

Modeled within the reactor building is the reactor pressure vessel and primary loop, including two steam generators, two reactor coolant pumps, and a pressurizer for the primary loop. It also contains industrial components including a fuel loading machine, polar crane, and a spent fuel cooling pool. The core contains the reactor fuel rods; inside the core is a mock-up fuel assembly with a 9$\times$9 array of fuel and control rods, which release heat through reactions or moderate reactions, respectively. The GPWR simulator uses a 17$\times$17 fuel assembly array, but the quarter-sized 9$\times$9 array is used in the model because it is simpler to model and less resource-intensive to render during simulations. The core is held by the reactor vessel, which has inlets for cold coolant to pumped by reactor coolant pumps and outlets that connects to hot leg and then to the steam generators. Coolant is pumped by the reactor coolant pumps into the reactor core, where it is heated up and enters into the steam generator to heat up the secondary loop. The pressurizer maintains pressure in the primary loop. The fuel loading machine handles the fuel assemblies for the loading and unloading operations. The spent fuel system is a special area for storing spent nuclear fuel assemblies that have been removed from the reactor core. The polar crane is used to lift and handle major components during various phases of a nuclear power plant's life cycle, including operation and decommissioning.

The turbine hall contains the power-generating components of the nuclear power plant to match those of the GPWR simulator: three low-pressure and one high-pressure turbines, an electric generator, and three condensers. The turbines are rotatory devices spun by steam in the secondary loop, effectively converting heat energy into mechanical energy. These turbines drives the electric generator, which converts mechanical energy into electrical energy. The moisture separator re-heaters are also modeled to improve the steam quality before enters into the low pressure turbine. As the steam flows through the turbines and becomes exhaust steam, it is then condensed into feedwater in the condensers. Feedwater then pumped through different stage of reheaters and then enters into the steam generator secondary side.

The final zone is the power area, where all of the electrical grid transmission systems are located. This contains an assemblage of high-voltage power lines and transformers that will carry electricity to the power grid, which ultimately supplies power to cities.

\section{Digital Twin Features}
The platform provides user-friendly features, enabling researchers to obtain accurate and detailed results, while easily accessing ongoing processes and outcomes in a single comprehensive view. These features include the Python-Unreal bridge, bidirectional real-time mirroring, robot models, virtual reality, robot navigation utilities, real-time top-down perspective, and thermal vision. Each feature is further explained in the subsequent sections.

\subsection{Python-Unreal Bridge}
The Python-Unreal bridge allows researchers to test their Python code within Unreal Engine. Since Unreal Engine is primarily based on C++, integrating and testing code can be challenging for researchers lacking a strong programming background. This feature helps to lower these barriers by enabling Python (a higher-level programming language more similar to human language) code to be executed within the engine. The UnrealCV plugin is utilized to allow for communication between the Python environment and Unreal Engine. This extension implements a client-server architecture, including a server that runs within the Unreal Engine and a Python client that allows sending commands to the server to manipulate actors and the environment. The plugin comes with predefined commands that the client can send, including setting and querying both actor and camera location and rotation. The plugin has been enhanced by developing additional commands, and both the server and client code have been extended to accommodate various test cases. For instance, code has been added to programmatically query a robot target navigation location placed on the map, to streamline the development and evaluation of target-based robot navigation tasks. 

While the Python-Unreal bridge provides a high-level form of communication with the Unreal environment, Python is also a popular language used for AI and ML. By integrating the ability to communicate with the Unreal Engine, iFANnpp opens up the potential for conducting various experiments while providing access to the entire Python AI/ML ecosystem.

\subsection{Bidirectional Real-Time Mirroring}

GPWR is a high-fidelity pressurized water reactor simulator that allows for real-time simulation of complex physical processes in nuclear power plants. In addition, the simulator incorporates physics-based models of fluid, gases, heat transfer, and other phenomena in nuclear power plants. Integrating the Unreal Engine project with the GPWR simulator enhances the framework by giving it a unique characteristic of a digital twin: continuous mirroring of real-time data, behavior, and interactions.

To enable the connection between the GPWR simulation software and iFANnpp environment, a Transmission Control Protocol (TCP) client and server architecture are implemented. The server part is a bridge between the GPWR server and the bidirectional pipe client. The client that runs within the Unreal Engine simulation makes requests to the server, which then relays commands to the GPWR simulation software. This procedure allows for reading and writing GPWR variables such as sensor pressure, water level, material weights, and liquid temperature. This connection is also used to update the sensor values within the GPWR environment. The client polls the server frequently for variable values, allowing for the corresponding values within the Unreal Engine simulation environment to synchronize with the GPWR variables as they change throughout the simulation. As the TCP client code is generic, the simulator could be replaced by a physical PWR system, using the same TCP client. Simulations can be run in real-time to make informed decisions about potential robot actions, and the best decision can be confirmed in simulation before deploying it to the real world. Since the iFANnpp platform synchronizes the state of its connected system with the simulation environment, the decisions made in simulation will be based on recent and relevant information.

The iFANnpp environment incorporates not only major 3D models but also interactive components, such as valves, pumps, and transducers. These components are capable of both reading data from and writing data to the digital twin and reflect changes in the GPWR simulation at every tick. Such changes can impact the entire nuclear power plant, triggering new updates based on the algorithm of the nuclear power plant's physics simulator. This process occurs continuously, with the digital twin regularly updated to reflect the real-world physics of the nuclear power plant. This bidirectional communication allows the digital twin to simulate real-time operational scenarios while setting simulator variables to change the scenario outcome, which is subsequently mirrored in the digital twin environment.

\subsection{Robot Models}

\begin{figure} [h!]
    \centering
    \includegraphics[width=\columnwidth]{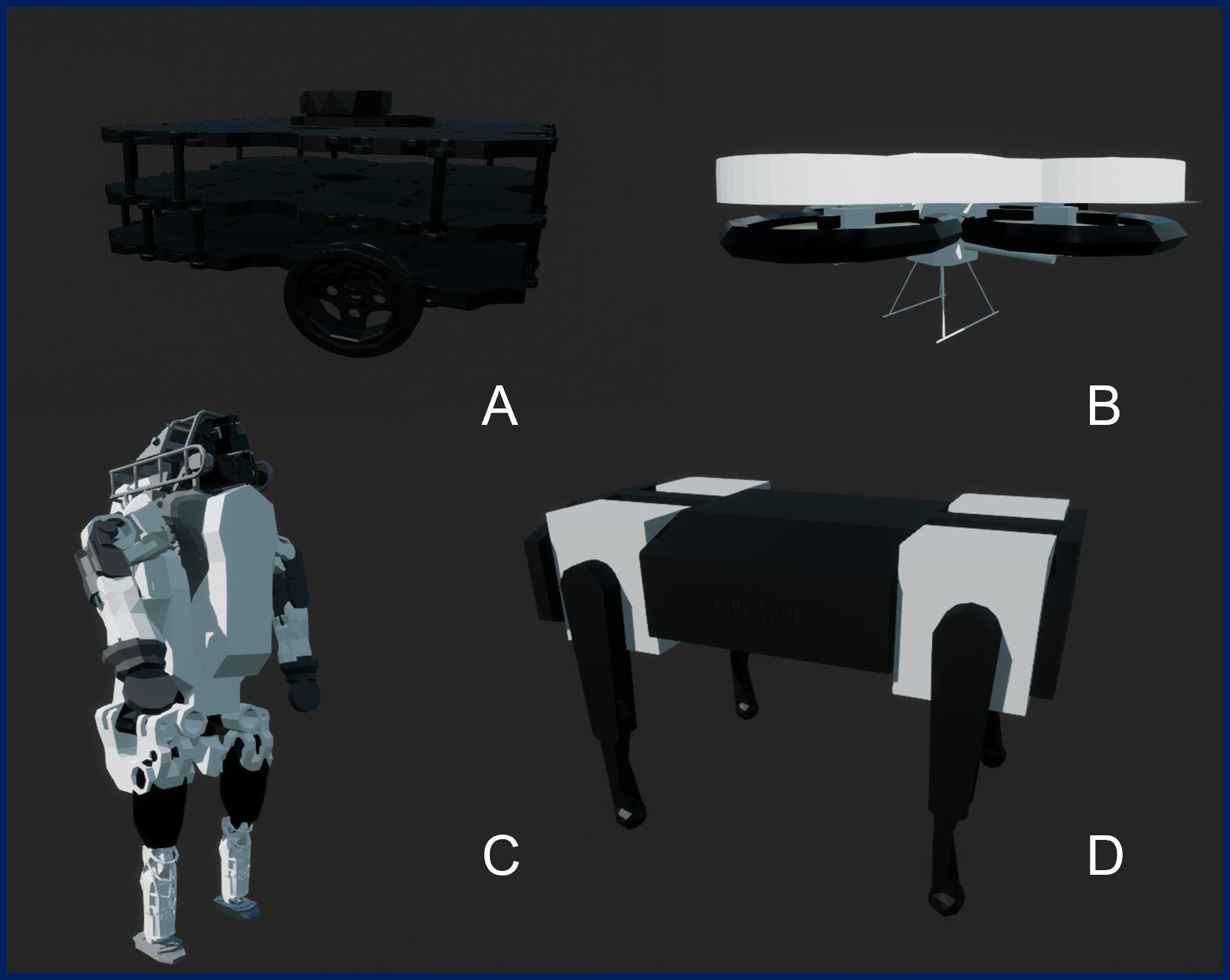}
    \caption{Robot models (A) wheeled (B) aerial (C) humanoid (D) quadruped}
    \label{fig:robot-models}
\end{figure}

A diverse range of robot models is provided within the digital platform to optimize operational capabilities across various tasks in nuclear power plants. Due to the different demands of tasks in such a highly complex and potentially hazardous environment, a variety of mobile systems are required. In iFANnpp, four distinct robot models with varying locomotion capabilities have been imported and developed within the digital twin environment: TurtleBot \cite{robotis_2022} (WheelRobot), Atlas \cite{virgir2015} (Bipedal Robot) , Champ \cite{Jimeno2020} (Quadruped Robot) , and Quadcopter \cite{MathWorks2017} (Aerial Robot).

Each of these robot types is tailored to specific functions. TurtleBot is used for tasks requiring high mobility on smooth, flat surfaces. It is ideal for routine tasks that require frequent movement across different plant sections on smooth surfaces. Atlas mimics human-like manipulator controls, allowing it to interact with human equipments. Champ is utilized in a situation where it requires stability in locomotion. The four-legged locomotion allows it to maintain balance on uneven terrain and provides agility and durability at the same time. Lastly, a quadcopter can inspect areas that are difficult to reach at altitudes. It can also be used to quickly survey or monitor large areas.

Use of robots within a digital twin can provide multiple potential benefits to the nuclear industry. First, robot models can be simulated and tested in design-specific tasks. The integration of robots with a digital twin allows for the customization and calibration of the robot models based on the specific needs of a given task. Second, developing and testing robot designs in a digital twin can reduce development time and cost. Researchers can experiment with various robot types, features, and workflows in a risk-free environment. Quick modifications can be made to the robots' locomotion, sensors, and operational algorithms without the need for expensive prototypes. Lastly, robots can be tested in potential hazardous areas without direct deployment. Nuclear power plants are highly sensitive environments where safety is critical. By deploying a variety of robots for specific tasks in a digital twin, the risk associated with such environment can be predicted and avoided.

\subsection{Virtual Reality}

Unreal Engine 5 provides powerful tools for VR development, enabling the design of immersive, interactive 3D environments. Leveraging these capabilities, a foundational VR implementation was developed to visualize iFANnpp, using Meta Quest 3 headset and controllers. VR-specific classes such as Pawn, Controller, and GameMode were created. The GameMode is designed to dynamically switch between different playing  modes (e.g.,  first  person, third person, robot arm perspective), triggered by a user-defined key. Core movement functionality was implemented, allowing for smooth navigation in two dimensions (forward, backward, left, and right) and rotational movement in the four directions. The feature also provides groundwork for further advancements such as integrating VR-based user interfaces and robotic control systems via VR.

The integration of VR into the digital twin offers several benefits, especially in the context of nuclear power plant simulation and training. First, VR enables users to engage in hands-on interaction within a fully immersive environment. This makes complex systems or components more straightforward to understand. This immersive experience accelerates training, especially when requiring extensive real-world setup. Second, VR minimizes the need to consider many variables associated with real-world tests, such as physical prototypes and on-site visits. Traditionally, system testing and training requires constructing expensive physical models or on-site presence. With VR, users can simulate real-world operations with high accuracy and reduce the costs associated with physical development. Lastly, changes or optimizations in systems or operations can be tested without risk of damaging actual systems.

\subsection{Robot Navigation Utilities}

\begin{figure}[t]
    \centering
    \includegraphics[width=\columnwidth]{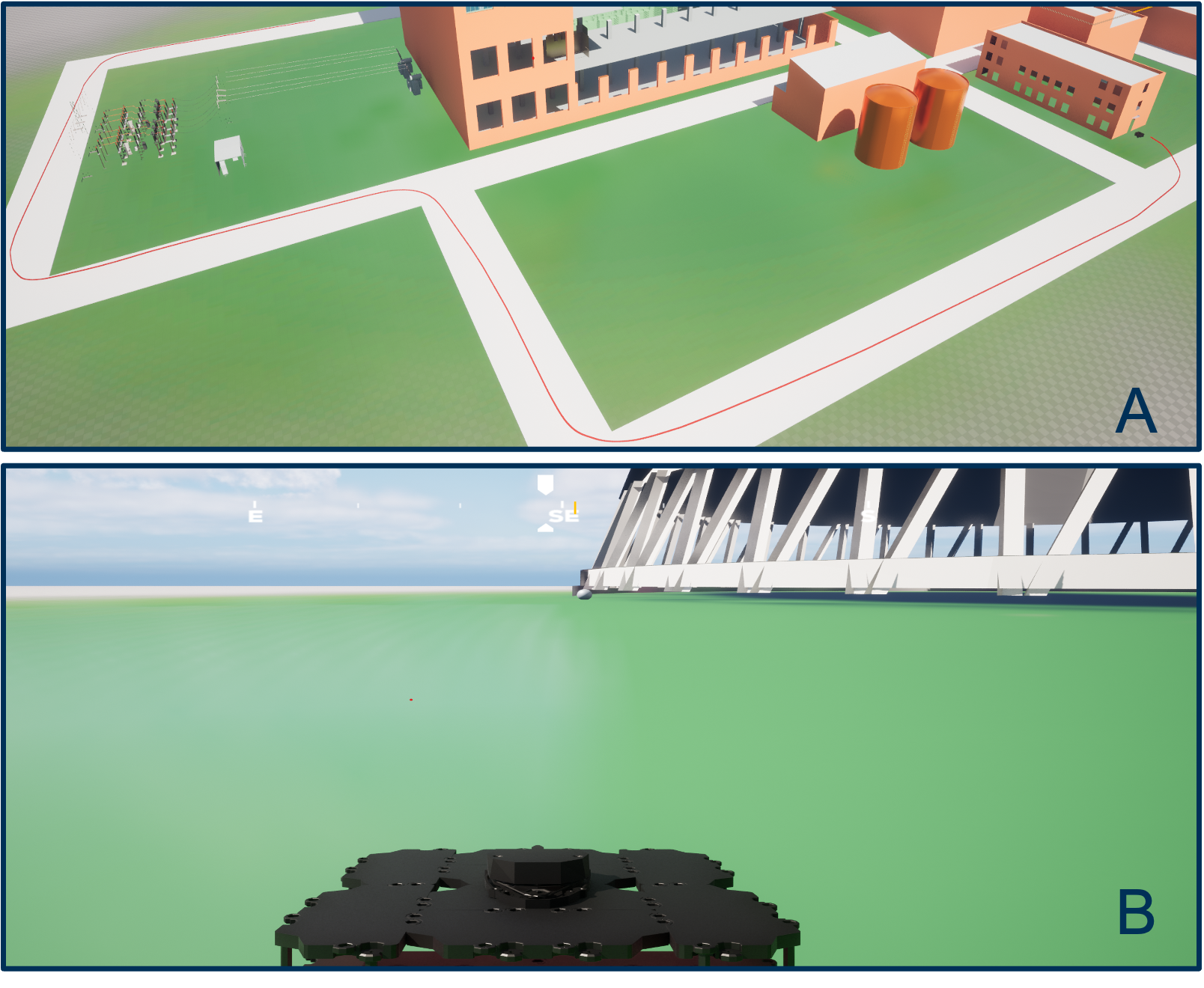}
    \caption{Navigation Utility}
    \label{fig:Navigation-Utility}
\end{figure}

iFANnpp includes functionality for visualizing the path that a robot takes. This feature is useful for robot debugging purposes, allowing researchers to visually compare a robot's expected path against the path it took in practice (See Figure \ref{fig:Navigation-Utility}). Researchers can also identify potential patterns in the path that the robot is taking, improving the interpretability of its learned algorithm and directing the researcher toward potential algorithmic or mechanical modeling modification to improve the robot's path planning and execution performance.

The path visualization feature is implemented in Unreal Engine as a component that can be added to a robot actor. A boolean variable can be selected through the editor to enable or disable the visualization feature. While enabled, the robot path is rendered onto the map. The robot path is implemented using a series of debug lines generated via calls to UE5 "DrawDebugLine". Successive points along the robot path are linearly interpolated to produce a smooth visual path that trails behind the robot. Both the thickness and color of the path can be customized through the editor.

Additionally, a boolean variable is exposed through the UE5 editor to allow for recording the robot location. The current robot location is recorded for every tick within the simulator and is logged to a CSV file along with the current timestamp in milliseconds. This logging feature can plot and visualize the robot path offline, independent of the iFANnpp environment, which can aid in post-hoc algorithm analysis. This feature can also be used to collect robot path demonstration trajectories to train robot agents, e.g., using behavioral cloning. Both human and machine policies can be used to produce demonstrations, as the robot can either be manually or pragmatically controlled. 

To enhance the interpretability of the robot path in the first-person perspective, a compass feature is developed. This feature can also aid researchers in reviewing and debugging first-person robot navigation task experimentation. The compass features can be enabled or disabled interactively by pressing a keyboard key while the digital twin simulation is running. This displays an overhead icon that tracks and highlights the target relative to the robot, indicating the relative angle of the target (see Figure \ref{fig:Navigation-Utility}). This can inform a practitioner on how to potentially adjust their algorithm to improve performance on a navigation task.

\begin{figure}[b]
    \centering
    \captionsetup{justification=centering}
    \includegraphics[width=\columnwidth]{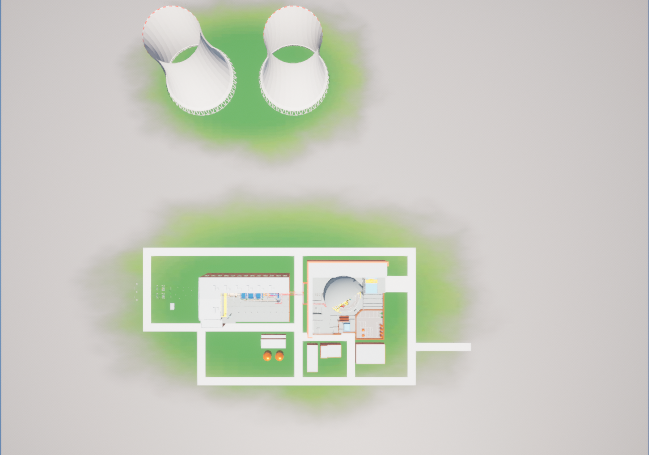}
    \caption{The top-down image of the entire digital twin}
    \label{fig:top-down feature}
\end{figure}

\begin{figure*}[ht]
    \centering
    \includegraphics[width=7in]{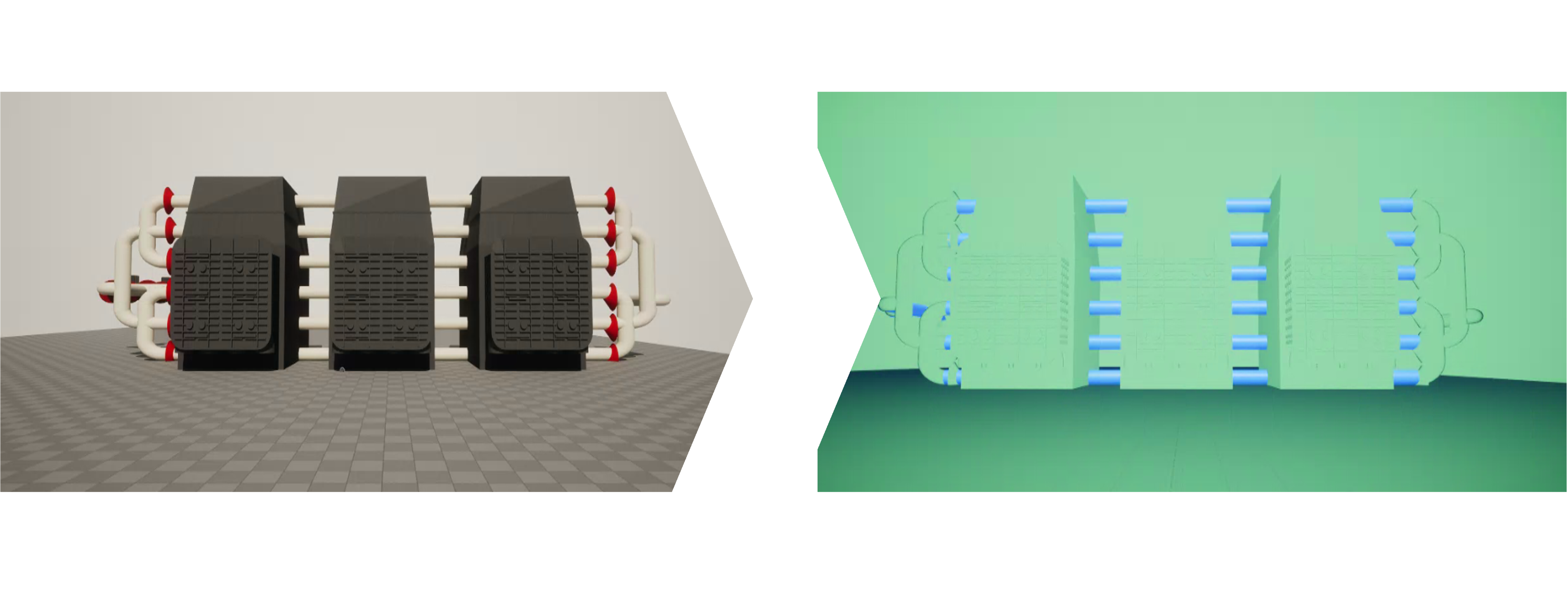}
    \caption{The thermal vision on the circulating water system}
    \label{fig:Thermal Vision}
\end{figure*}

\subsection{Real-Time Top-Down Perspective}

In the context of robot navigation within this three-dimensional environment, visualizing a robot path and understanding its actions due to underlying mechanical and algorithmic nuances can be challenging. To mitigate these difficulties, a real-time top-down perspective feature was developed.

A top-down camera class was created within Unreal Engine, inheriting from the camera actor class. This camera is positioned at a location in which the entire map can be captured. It becomes active through a single key input. This feature allows switching between first-person and top-down perspectives. The change in perspective is useful as it provides a comprehensive overview of the environment, shown in Figure \ref{fig:top-down feature}. The top-down images are saved at predefined intervals to record critical moments and movements.

This feature offers several benefits. First, the top-down feature provides tracing the robot's path. This perspective displays a clear visualization of the route taken by the robot, which helps researchers to understand the robot's movement patterns. Second, this feature enables the precise identification of multiple robots that work simultaneously in the environment. In scenarios where coordination and interaction between several robots are critical, this feature can be utilized effectively. Third, the feature allows for comprehensive process monitoring of the entire nuclear power plant. Researchers can view the environment from above and track the interactions between various elements. Lastly, the feature captures a series of images throughout an operation at predefined intervals. The saving of images establishes a continuous record of the robot's activities, which can be used for operation analysis, troubleshooting, and performance optimization.

\subsection{Thermal Vision}

iFANnpp is integrated with GPWR and continuously updates various parameters in real-time. Access to real-time temperature data can lead to numerous research opportunities, such as maintenance task scheduling of temperature-based systems (e.g., cooling water system), routine inspection tasks, and robot path optimization (e.g., directing a robot around high-temperature areas to avoid damage). A thermal camera integrates both visual and thermal information. The deployment of a robotic system equipped with thermal vision within a nuclear power plant can enable the implementation of navigation control that requires both information modalities. An advantage of thermal vision over manual sensor inspection is that it can be utilized at a distance, while manual measurement operation may require close physical proximity to a sensor. In addition, while wireless communication can facilitate the transfer of temperature-based information \cite{gungor2009industrial}, thermal vision serves as an additional avenue for encoding temperature. This method can improve the robustness of the system when there is a possibility of wireless interference. Finally, thermal vision can also effectively serve as a substitute mobile temperature sensor if stationary temperature sensors malfunction.

A thermal vision feature was implemented using "Material" function from Unreal Engine. Thermal values are linearly interpolated between $-100$ to $100$ Celsius, visually represented by interpolating from blue to red. Actors without temperature properties are visually distinguished by being rendered in green, effectively removing them from the blue and red spectrum. This color exclusion process eliminates confusion with Unreal Engine actors that are naturally red or blue. As shown in Figure \ref{fig:Thermal Vision}, this method provides a clear thermal view of the circulating water system, allowing users to interpret the temperature distribution easily. With real-time mirroring, this feature dynamically reflects the constantly changing environment and displays the entire map in thermal view.

\begin{figure*}[t]
  \centering
  {\includegraphics[width=7in]{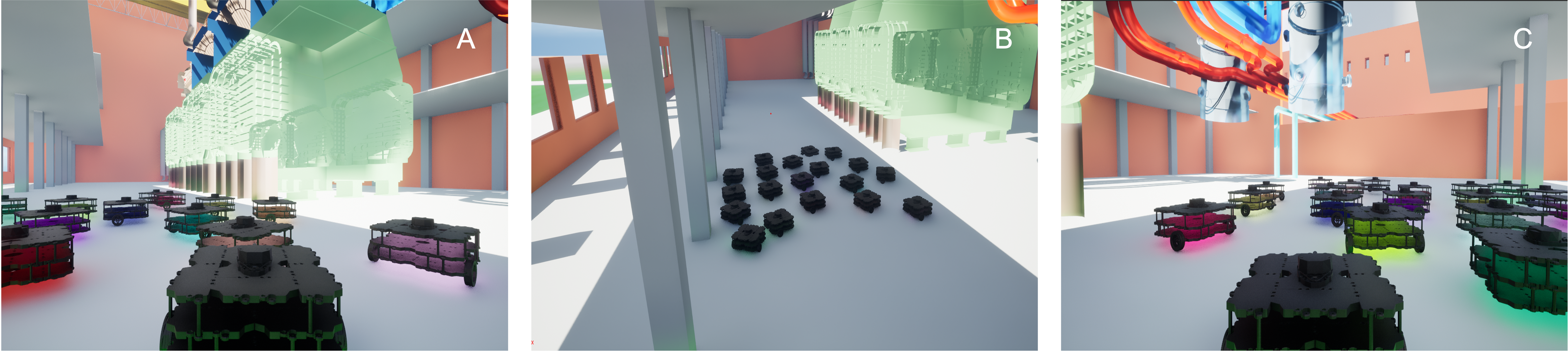}}
  \caption{Robotic swarm simulation. (a) and (c) display first-person individual robot views. (b) displays the entire swarm, contained within the turbine hall.}
  \label{fig:robot-swarm}
\end{figure*}

This feature has many potential applications. First, the thermal view provides an intuitive and immediate understanding of temperature distributions across the plant. As mentioned, robots can make quicker decisions without the need to physically walk up to individual sensors. Second, this feature can enable comprehensive plant monitoring, ensuring that all critical components are consistently observed. Third, thermal images can be synthetically generated to address the lack of open-source visual thermal data in nuclear power plants. Moreover, the generated thermal data can uncover patterns, trends, and correlations that are difficult to observe in tabular datasets. While tabular data often require thorough analysis to identify subtle trends, this feature captures complex and minor variations in the environment, enabling the perception of multiple nuanced patterns simultaneously.

\section{Use Cases}

In this section, a representative set of realistic scenarios is provided, utilizing iFANnpp for future research projects. These scenarios demonstrate multiple research directions that could take advantage of the features provided by the iFANnpp platform.

\subsection{Example Scenario: Swarm of Robots}

One valuable avenue for robotics research in nuclear power plants (as well as industrial systems in general) is swarm robotics \cite{alremeithi2024use, ikumapayi2024swarm}. Robotics swarms allow for the orchestration of a relatively large number of simplistic robots capable of coordinating together to perform complex behavior. Robotic swarm deployment can be used to form flexible functional configurations, create mobile assembly lines, and perform coordinated maintenance and repair tasks. 

To demonstrate the viability of simulating a robotic swarm, a swarm of twenty TurtleBots was placed within the turbine hall of the iFANnpp environment. The robots were placed adjacent to one another in a closely clustered configuration. Figure \ref{fig:robot-swarm} displays the swarm robot simulation. The left-most and right-most images display first-person camera views from two individual robots within the swarm. Each robot was individually controlled from these views. The middle image is an overhead display of the twenty-bot swarm. By setting net mode within the Unreal Editor to "Play as Listen Server" and extending the number of players to 2, multiple turtle-bots were controlled using the mouse and keyboard. Control was transferred from one robot to another. As with other simulation tasks, iFANnpp allows for iterating on both the mechanical and algorithmic components of swarm robot systems. 

While the demonstration involved manual robot control, this paradigm can be extended to control multiple robots in a near-parallel fashion programmatically. Although simplistic, this scenario opens the possibilities for developing and refining multi-agent algorithms intended for deployment within nuclear power plants. 

\subsection{Example Scenario: Robot Remote Control Via Digital Twin}

\begin{figure}[b]
    \centering
    \includegraphics[width=\columnwidth]{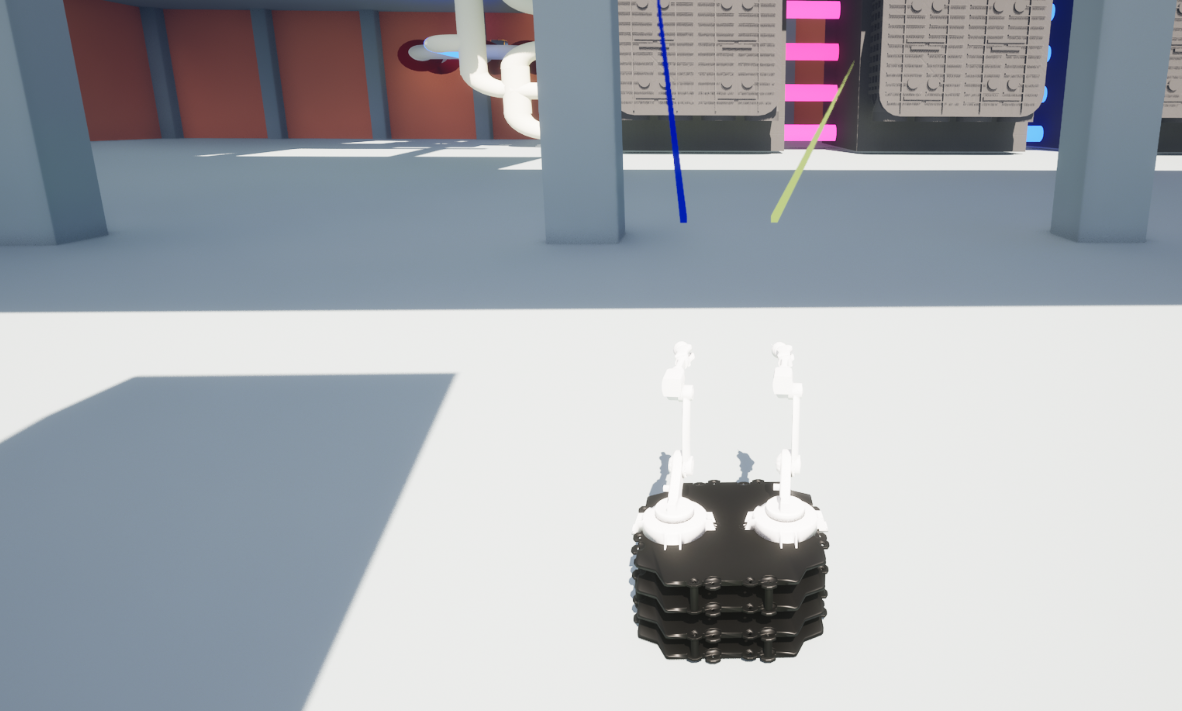}
    \caption{Robotic teleportation simulation.}
    \label{fig:robot-teleoperation}
\end{figure}

The implementation of VR technology within the digital twin, especially in critical infrastructure like nuclear power plants, provides significant value. Teleoperation, remotely control of robots using the VR technology, is a key method for ensuring safe operation. In the scenario illustrated in Figure \ref{fig:robot-teleoperation}, a teleoperation setup is demonstrated in the turbine hall. The VR allows an operator to control a robot remotely and navigate the robot to locate and close a valve to prevent the system from overflooding. 

This application demonstrates how VR can increase safety, enabling operators to address emergencies without being physically present in dangerous areas. Furthermore, VR improves operator remote working accuracy and reduced cognitive load through enhanced situational awareness.

There are multiple research directions that can utilize the integration of VR technology with nuclear power plants. Scenarios such as mobile robot teleoperation for pipe maintenance \cite{Do2024teleoperation} and training platform development \cite{Do2024VRPlatform} have been experimentally validated, and the corresponding results are reported in the references cited. Other topics, including advanced human-robot collaboration and disaster preparedness simulations, remain conceptual at this stage and are subjects of ongoing and planned research.

\subsection{Example Scenario: Robot Navigation}

Research has focused extensively on enhancing the use of robots in industrial systems, including work focusing on their movement and decision-making capabilities \cite{hesu2025preference}. One critical feature is their ability to navigate effectively, which involves selecting optimal paths while avoiding hazardous areas, such as those with high radiation levels in nuclear power plants. While robot teleportation allows for manual human control, autonomous robot navigation can be valuable when a human is not present or able to operate a robot remotely (e.g., if it is difficult to establish wireless communication). In addition, autonomous navigation capabilities can remove the burden on human operators for mundane mobile tasks such as fetching a tool that the operator needs. For example, robots are utilized in autonomous rapid radiation reconstruction in the work \cite{son2025physics}, which could be tested in iFANnpp. Therefore, research investigating the design and improvement of autonomous robot navigation algorithms is of interest.

Integration of the Python-Unreal bridge with iFANnpp enables the development of novel robot navigation algorithms, critical for mobile robot deployment within an NPP. State information, such as location and RGB camera images, can be queried from the environment and used to inform a robot policy to perform intelligent navigation, including reinforcement-learning-driven control. With this workflow, many scenarios can be evaluated, such as physically constrained navigation with unanticipated environmental changes.

\subsection{Example Scenario: Cyber-Physical Security Testing}

With the rapid digitalization and emerging technology adoption in nuclear field, cybersecurity requires more attention and further research \cite{fancyberbook}. Previous works have employed the use of the GPWR simulator for cybersecurity testing. Yoo et al. \cite{yoo2025self} utilized GPWR to simulate a false-data-injection attack in order to design an automated attack mitigation strategy using autoencoder output to correct for the attack effects. Chen et al. \cite{chen2024full} used GPWR as the backbone for hardware-in-the-loop testing and conducted a supply chain attack simulation for future cyber-attack development and NPP system robustness testing. While the data from this GPWR can be used to test different cyber-attack scenarios, integration of the simulator with the physically modeled iFANnpp plant allows for additionally incorporating physical information with security-focused experiments. This enables the development of multimodal cyber-attack prevention or detection techniques - both physical information (such as from images), as well as numerical plant sensor values can be used to identify potential threats to the plant to prevent damage. 

Example scenarios can vary by testing different degrees of physical and cyber-related consequences and effects. Physical tracking can be used to identify if an unauthorized entity is located within the plant. Correlation of process data with observed robot or human function can help to identify when physical interaction with the plant is benign or malicious. 

While humans present one physical attack vector, it is also critical to ensure the security of the robots themselves for practical real-world robot deployment for monitoring and maintenance tasks within an NPP. Additional tests can be conducted such as identifying if a robot is being maliciously controlled by combining technologies such as action identification and anomaly detection with plant process data. Unintentional robot malfunction identification technique development can also be conducted.

\section{Profiling}

The main objective of the development of iFANnpp is to provide a platform for researchers to investigate various robotics research in nuclear power plants quickly and easily. Validating and demonstrating the effectiveness, efficiency, and reliability of the platform is essential. To achieve this, a series of profiling tests were conducted to measure the runtime performance of various platform operations, along with memory and CPU usage. 

Profile tests were performed on a single Intel(R) Xeon(R) w7-2475X machine, containing 20 cores and 40 threads. Attached to this machine were two NVIDIA RTX A6000 GPUs. Note that the UnrealEditor process was only using memory from one GPU, per the NVIDIA System Management Interface (nvidia-smi) utility. The following features were profiled: GPWR communication, thermal vision, top-down perspective capture, VR movement, and multiple commands using the Python-Unreal bridge.

\begin{table*}[t]
    \renewcommand{\arraystretch}{1.5} 
    \centering
    \begin{tabularx}{\textwidth}{|c||X|X|X|X|X|}
        \hline
        \multicolumn{1}{|c||}{} & \multicolumn{1}{c|}{\textbf{Speed Test}} & \multicolumn{2}{c|}{\textbf{Memory Test}} & \multicolumn{1}{c|}{\textbf{Functional Test}} \\ \hline 
        \multicolumn{1}{|c||}{} & \textbf{Time} & \textbf{Memory Usage} & \textbf{CPU Usage} & \textbf{Unit Test} \\ \hline\hline
        \textbf{Baseline} & - & 3.37 GB  & 108.76\% & - \\ \hline
        \textbf{GPWR Communication(set)} & 18.16 ms & 3.72 GB  & 85.52\% & Passed (100\%) \\ \hline
        \textbf{GPWR Communication(get)} & 2.37 ms &  3.63 GB & 426.40\% & Passed (100\%) \\ \hline
        \textbf{Unreal-Python Bridge (Read Image)} & 27.79 ms  & 4.56 GB & 235.52\% & Passed (100\%) \\ \hline 
        \textbf{Unreal-Python Bridge (Environment Step)} &  56.70 ms  & 4.59 GB & 255.58\% & Passed (100\%) \\ \hline 
        \textbf{Unreal-Python Bridge (Restart Episode)} &  104.98 ms  & 4.61 GB & 255.07\% & Passed (100\%) \\ \hline 
        \textbf{Unreal-Python Bridge (Move 2D)} &  20.93 ms  & 4.60 GB & 233.87\% & Passed (100\%) \\ \hline
        \textbf{Thermal Vision} & 87.62 ms & 4.42 GB &  186.06\% & Passed (100\%)  \\ \hline 
        \textbf{Real-Time Top-Down Perspective} & 690.55 ms  & 3.66 GB & 76.87\% & Passed (100\%)  \\ \hline 
        \textbf{Virtual Reality} & 720.47 ms & 4.45 GB & 273.56 \% & Passed (100\%)  \\ \hline 
    \end{tabularx}
    \caption{Performance Metrics}
    \label{tab:Performance-Metrics}
\end{table*}

\subsection{Speed Tests}

A speed test was conducted to evaluate the time required to perform individual operations, measured in milliseconds. Table \ref{tab:Performance-Metrics} presents the execution time for each feature in its first column.

To ensure precise measurement of operation runtime, the current timestamp was logged in milliseconds using the C++ standard chrono library, both before and after the operation. This approach, implemented via a custom logger actor component within the Unreal Engine blueprint, allowed the runtime for each operation to be calculated as the difference between the two timestamps. This method was applied to all operations except for the Python-Unreal bridge navigation commands. For these commands, timestamps were recorded before and after sending commands through the unrealcv Python client, utilizing Python's time library.

Separate profiling was performed for the read and write operations as part of the GPWR communication test. The read operation queried the current state of the GPWR variables, while the write operation updated the state of the GPWR variables. For both tests, 100 GPWR variables were transmitted between the iFANnpp client and the GPWR server. Each read and write operation was performed 100 times, with the execution time recorded in milliseconds. This test quantified the time required for the iFANnpp platform to synchronize the variables with GPWR. The average execution times for read and write operations are detailed in the table. The write operation yielded an average runtime of 18.16 milliseconds the read operation averaged 2.37 milliseconds.  

The Python-Unreal bridge navigation commands test measured the runtime for the following independent operations: (1) capturing a first-person (robot perspective) image from iFANnpp via the Python client, (2) moving the robot in two dimensions (e.g., turning left or right by 15°, or moving forward or backward by 1 meter), with actions chosen randomly, and (3) restarting the Unreal Engine level or training episode. Additionally, the total time to complete a single "step" operation was profiled, as defined by the Gymnasium v0.29.1 API \cite{DBLP:journals/corr/BrockmanCPSSTZ16}. The "step" function involved moving the robot, capturing a first-person image, and calculating the reward. Each navigation command operation was executed 100 times, and the average execution time is detailed in the table. These operations were profiled as they can be used as fundamental operations for the purposes of reinforcement-learning-based navigation. The most expensive operation was restarting the episode, which took 104.977 milliseconds on average, while reading a first-person image and moving one step averaged 27.79 and 20.93 milliseconds, respectively. The total time cost per environment step was 56.7 milliseconds. 

The thermal vision profile test measured the time required to visually update a component of the environment with thermal information, excluding the time taken to retrieve this information from GPWR. Thermal updates were manually triggered using a pre-programmed key press, and the total time for each update operation was recorded. The average time was calculated over 100 updates and is reported in the table, which took an average of 87.62 milliseconds per operation. 

The real-time top-down test evaluated the time needed to capture a sequence of top-down images at fixed time intervals. A single operation was performed to capture 100 consecutive images, with a one-second delay between captures. The average time across 100 images is recorded as the final value in the table. The one-second delays were subtracted from the total capture times to isolate the operational overhead. The average time per 100 recorded images was 690.55 milliseconds. Note that this operation can be performed concurrently with other operations such as robot navigation.

The VR performance test measured the time required to execute a move command in a VR environment, covering both the time taken to send the command from the client to the server and the time to receive the corresponding response. Each move command instructed the VR entity to navigate within the simulated environment, performing a basic action to move forward or backward by a specified distance. A total of 100 move commands were executed and the execution time for each command was recorded in milliseconds. The average execution time across all 100 commands yielded a runtime of 720.47 millisecond average.

The results of the speed test demonstrate that the digital twin is applicable to multiple research directions. The relatively low latency of the platform indicates that several features can be used without critical delays. This is particularly important for AI model training, which may require substantial computational resources on top of the iFANnpp overhead. The results therefore signify that the platform is suitable for model training as well.

\subsection{Memory Tests}

Memory and CPU usage were measured consistently across all operations described in the previous section using a standardized protocol. Memory usage tracked the platform's memory consumption under varying load conditions, while CPU usage was analyzed to evaluate the utilization of processing resources. CPU usage, reported as a percentage, indicates the proportion of cores in use; values exceeding 100\% signify the utilization of multiple cores (e.g., 200\% CPU usage corresponds to two fully utilized CPU cores).

The Linux top command was used to monitor the memory and CPU usage for each operation. This command was run continuously for 60 seconds, covering the duration during which all profile runs for a specific operation were executed. Instantaneous memory and CPU usage were recorded every second. Both the mean and maximum values for memory and CPU usage were recorded. The memory and CPU usage metrics for each experiment are summarized in the second and third columns of Table \ref{tab:Performance-Metrics}.

To establish a baseline, the usage of memory and CPU was also measured during the idle state of iFANnpp. In this scenario, no additional operations were performed, providing a reference for comparison with the resource consumption of other tests.

The baseline test consumed roughly 3.37GB of RAM with \%108.75 CPU usage. No operation exceeded 5GB of RAM or the use of 5 CPU cores. Therefore the resources required to perform these core feature operations with iFANnpp are accessible to the average researcher.

\subsection{Functional Tests}

For functional testing, unit tests were conducted for each operation to ensure that all developed features functioned as intended. Each unit test verified the correctness and stability of an individual feature. The task for each feature is described in the previous section and tested accordingly. The unit test for each feature starts by launching the project in UE5. A task operation was executed ten times consecutively. The project was closed and reopened, and the operation was executed ten more times. This cycle was repeated until the operation reached a total of 100 times.

As shown in Table \ref{tab:Performance-Metrics}, all tests were successfully passed. The result demonstrates the robustness and reliability of the digital twin and its features implemented. The perfect result across repeated executions underscores the correctness of the implementations and ensures their readiness for practical application in research settings.

\section{Conclusion}
iFANnpp, a digital twin for nuclear power plants for robots and autonomous intelligence, models a comprehensive nuclear power plant with real-time bidirectional updates with GPWR, the high-fidelity physics simulator for a nuclear power plant. This setup enables researchers to validate robot planning and control methods at a larger scale. The digital twin models the major physical components of a PWR, such as the turbine hall, cooling towers, water sources, and power output area. A Python-AI bridge is provided to enable robotics and related research areas, incorporating user-friendly access to a range of AI libraries and easier accessibility for users.

The digital twin offers features for designing and testing robot algorithms within nuclear power plants. It supports multiple mobile robots, including wheeled, bipedal, quadruped, and aerial robots. Additionally, the digital twin incorporates the VR feature with controller support, providing users with an immersive environment. For major areas of robotics research, such as task planning and perception, this digital twin also integrates features like navigation utilities, thermal vision, and real-time top-down perspectives.

Various experimental scenarios were conducted to validate the platform's functionality and reliability. A swarm of robots was created and manually controlled to explore possibilities for multi-robot research. Teleoperation using VR was also investigated, demonstrating the system's ability to enhance realism in robot control and interaction with nuclear power plant components. Core robot navigation operations were validated, which can be used for RL/AI applications. Additionally, the digital twin underwent comprehensive profiling tests to assess runtime performance. Speed tests, memory usage tests, and unit tests were performed on various platform features, confirming its reliability and effectiveness for robotics research.

Several approaches can be explored to extend this research further. First, the platform can be modeled with different industrial environments, such as smart factories and manufacturing plants. Many of the features of the platform can be applied not only to the nuclear power plant operations but also to a wide range of industrial tasks. Second, the challenges of nuclear power plants can be addressed using advanced AI algorithms, including RL techniques. The Python bridge enables rapid and user-friendly testing and validation of these methods. Finally, additional features relevant to nuclear power plant environments, such as radiation vision and enhanced locomotion capabilities to navigate diverse terrains, can be developed to expand the functionality of the platform further.

iFANnpp holds significant potential for advancing robotics research in nuclear power plants. Its integration of realistic simulations, diverse features, and AI model training capabilities opens the way for the future of autonomous systems. The platform’s reliability establishes it as a robust testbed, supporting further exploration and development in robotics research within the nuclear industry.

\section*{ACKNOWLEDGMENT}

The authors thank the U.S. Department of Energy, Office of Nuclear Energy, Distinguished Early Career Program for funding this work, award number DE-NE0009306.

\ifCLASSOPTIONcaptionsoff
  \newpage
\fi


\bibliographystyle{ieeetr}
\bibliography{iFANnpp}

\begin{thebibliography}{10}

\bibitem{von2021precise}
F.~von Drigalski, K.~Hayashi, Y.~Huang, R.~Yonetani, M.~Hamaya, K.~Tanaka, and Y.~Ijiri, ``Precise multi-modal in-hand pose estimation using low-precision sensors for robotic assembly,'' in {\em 2021 IEEE International Conference on Robotics and Automation (ICRA)}, pp.~968--974, IEEE, 2021.

\bibitem{katsamenis2022simultaneous}
I.~Katsamenis, N.~Doulamis, A.~Doulamis, E.~Protopapadakis, and A.~Voulodimos, ``Simultaneous precise localization and classification of metal rust defects for robotic-driven maintenance and prefabrication using residual attention u-net,'' {\em Automation in Construction}, vol.~137, p.~104182, 2022.

\bibitem{pretto2020building}
A.~Pretto, S.~Aravecchia, W.~Burgard, N.~Chebrolu, C.~Dornhege, T.~Falck, F.~Fleckenstein, A.~Fontenla, M.~Imperoli, R.~Khanna, {\em et~al.}, ``Building an aerial--ground robotics system for precision farming: an adaptable solution,'' {\em IEEE Robotics \& Automation Magazine}, vol.~28, no.~3, pp.~29--49, 2020.

\bibitem{wu2022human}
M.~Wu, J.-R. Lin, and X.-H. Zhang, ``How human-robot collaboration impacts construction productivity: An agent-based multi-fidelity modeling approach,'' {\em Advanced Engineering Informatics}, vol.~52, p.~101589, 2022.

\bibitem{lagomarsino2022robot}
M.~Lagomarsino, M.~Lorenzini, E.~De~Momi, and A.~Ajoudani, ``Robot trajectory adaptation to optimise the trade-off between human cognitive ergonomics and workplace productivity in collaborative tasks,'' in {\em 2022 IEEE/RSJ International Conference on Intelligent Robots and Systems (IROS)}, pp.~663--669, IEEE, 2022.

\bibitem{johnson2022multi}
D.~Johnson, G.~Chen, and Y.~Lu, ``Multi-agent reinforcement learning for real-time dynamic production scheduling in a robot assembly cell,'' {\em IEEE Robotics and Automation Letters}, vol.~7, no.~3, pp.~7684--7691, 2022.

\bibitem{truong2021backstepping}
T.~N. Truong, A.~T. Vo, and H.-J. Kang, ``A backstepping global fast terminal sliding mode control for trajectory tracking control of industrial robotic manipulators,'' {\em IEEE Access}, vol.~9, pp.~31921--31931, 2021.

\bibitem{he2020admittance}
W.~He, C.~Xue, X.~Yu, Z.~Li, and C.~Yang, ``Admittance-based controller design for physical human--robot interaction in the constrained task space,'' {\em IEEE Transactions on Automation Science and Engineering}, vol.~17, no.~4, pp.~1937--1949, 2020.

\bibitem{wang2022control}
J.~Wang and A.~Chortos, ``Control strategies for soft robot systems,'' {\em Advanced Intelligent Systems}, vol.~4, no.~5, p.~2100165, 2022.

\bibitem{li2023design}
X.~Li, H.~Yu, H.~Feng, S.~Zhang, and Y.~Fu, ``Design and control for wlr-3p: a hydraulic wheel-legged robot,'' {\em Cyborg and Bionic Systems}, vol.~4, p.~0025, 2023.

\bibitem{taheri2023study}
H.~Taheri and N.~Mozayani, ``A study on quadruped mobile robots,'' {\em Mechanism and Machine Theory}, vol.~190, p.~105448, 2023.

\bibitem{li2023aerial}
Q.~Li, H.~Li, H.~Shen, Y.~Yu, H.~He, X.~Feng, Y.~Sun, Z.~Mao, G.~Chen, Z.~Tian, {\em et~al.}, ``An aerial--wall robotic insect that can land, climb, and take off from vertical surfaces,'' {\em Research}, vol.~6, p.~0144, 2023.

\bibitem{qiu2017unrealcv}
W.~Qiu, F.~Zhong, Y.~Zhang, S.~Qiao, Z.~Xiao, T.~S. Kim, and Y.~Wang, ``Unrealcv: Virtual worlds for computer vision,'' in {\em Proceedings of the 25th ACM international conference on Multimedia}, pp.~1221--1224, 2017.

\bibitem{sorensen2022potentials}
J.~V. S{\o}rensen, Z.~Ma, and B.~N. J{\o}rgensen, ``Potentials of game engines for wind power digital twin development: an investigation of the unreal engine,'' {\em Energy Informatics}, vol.~5, no.~Suppl 4, p.~39, 2022.

\bibitem{garg2021digital}
G.~Garg, V.~Kuts, and G.~Anbarjafari, ``Digital twin for fanuc robots: Industrial robot programming and simulation using virtual reality,'' {\em Sustainability}, vol.~13, no.~18, p.~10336, 2021.

\bibitem{erdei2022design}
T.~I. Erdei, R.~Krak{\'o}, and G.~Husi, ``Design of a digital twin training centre for an industrial robot arm,'' {\em Applied Sciences}, vol.~12, no.~17, p.~8862, 2022.

\bibitem{metzner2020system}
M.~Metzner, D.~Utsch, M.~Walter, C.~Hofstetter, C.~Ramer, A.~Blank, and J.~Franke, ``A system for human-in-the-loop simulation of industrial collaborative robot applications,'' in {\em 2020 IEEE 16th International Conference on Automation Science and Engineering (CASE)}, pp.~1520--1525, IEEE, 2020.

\bibitem{gonzalez2024approach}
R.~Gonz{\'a}lez-Herb{\'o}n, G.~Gonz{\'a}lez-Mateos, J.~R. Rodr{\'\i}guez-Ossorio, M.~Dom{\'\i}nguez, S.~Alonso, and J.~J. Fuertes, ``An approach to develop digital twins in industry,'' {\em Sensors}, vol.~24, no.~3, p.~998, 2024.

\bibitem{vairagade2024nuclear}
H.~Vairagade, S.~Kim, H.~Son, and F.~Zhang, ``A nuclear power plant digital twin for developing robot navigation and interaction,'' {\em Frontiers in Energy Research}, vol.~12, p.~1356624, 2024.

\bibitem{hoebert2023knowledge}
T.~Hoebert, W.~Lepuschitz, M.~Vincze, and M.~Merdan, ``Knowledge-driven framework for industrial robotic systems,'' {\em Journal of Intelligent Manufacturing}, vol.~34, no.~2, pp.~771--788, 2023.

\bibitem{ipiales2023virtual}
J.~S. Ipiales, E.~J. Araque, V.~H. Andaluz, and C.~A. Naranjo, ``Virtual training system for the teaching-learning process in the area of industrial robotics,'' {\em Electronics}, vol.~12, no.~4, p.~974, 2023.

\bibitem{bilancia2023overview}
P.~Bilancia, J.~Schmidt, R.~Raffaeli, M.~Peruzzini, and M.~Pellicciari, ``An overview of industrial robots control and programming approaches,'' {\em Applied Sciences}, vol.~13, no.~4, p.~2582, 2023.

\bibitem{farooq2023power}
M.~U. Farooq, A.~Eizad, and H.-K. Bae, ``Power solutions for autonomous mobile robots: A survey,'' {\em Robotics and Autonomous Systems}, vol.~159, p.~104285, 2023.

\bibitem{naveen2023wireless}
S.~Naveen, K.~Prasad, D.~Gowda, A.~Ranjan, K.~K. Hegge, and R.~Agrawal, ``Wireless and solar-powered multipurpose robot for industrial automation: A sustainable solution,'' in {\em 2023 8th International Conference on Communication and Electronics Systems (ICCES)}, pp.~146--151, IEEE, 2023.

\bibitem{ghodsian2023mobile}
N.~Ghodsian, K.~Benfriha, A.~Olabi, V.~Gopinath, and A.~Arnou, ``Mobile manipulators in industry 4.0: A review of developments for industrial applications,'' {\em Sensors}, vol.~23, no.~19, p.~8026, 2023.

\bibitem{buerkle2023towards}
A.~Buerkle, W.~Eaton, A.~Al-Yacoub, M.~Zimmer, P.~Kinnell, M.~Henshaw, M.~Coombes, W.-H. Chen, and N.~Lohse, ``Towards industrial robots as a service (iraas): Flexibility, usability, safety and business models,'' {\em Robotics and Computer-Integrated Manufacturing}, vol.~81, p.~102484, 2023.

\bibitem{EnergyEncyclopedia}
EnergyEncyclopedia, ``Nuclear power plant pwr 3d models,'' 2024.

\bibitem{gpwr}
WSC, ``Wsc inc., nuclear power plant simulation.'' \url{https://www.ws-corp.com/default.asp?PageID=2\&PageNavigation=Nuclear-Power-Plant-Simulation}, 1995.

\bibitem{robotis_2022}
Robotis, ``turtlebot3,'' 2022.
\newblock Github repository: https://github.com/ROBOTIS-GIT/turtlebot3.

\bibitem{virgir2015}
Vigir, ``Virgiratlascommon,'' 2015.
\newblock Github repository: https://github.com/team-vigir.

\bibitem{Jimeno2020}
J.~M. Jimeno, ``champ,'' 2020.
\newblock Github repository: https://github.com/chvmp/champ.

\bibitem{MathWorks2017}
MathWorks, ``quadcopter-simulation-ros-gazebo,'' 2017.
\newblock GitHub repository: https://github.com/mathworks/quadcopter-simulation-ros-gazebo.

\bibitem{gungor2009industrial}
V.~C. Gungor and G.~P. Hancke, ``Industrial wireless sensor networks: Challenges, design principles, and technical approaches,'' {\em IEEE Transactions on industrial electronics}, vol.~56, no.~10, pp.~4258--4265, 2009.

\bibitem{alremeithi2024use}
K.~Alremeithi and W.~Sealy, ``The use of digital twin for mobile robot swarm task allocation,'' {\em Manufacturing Letters}, vol.~41, pp.~1200--1208, 2024.

\bibitem{ikumapayi2024swarm}
O.~M. Ikumapayi, O.~T. Laseinde, R.~R. Elewa, T.~S. Ogedengbe, and E.~T. Akinlabi, ``Swarm robotics in a sustainable warehouse automation: Opportunities, challenges and solutions,'' in {\em E3S Web of Conferences}, vol.~552, p.~01080, EDP Sciences, 2024.

\bibitem{Do2024teleoperation}
Y.~Do, C.-Y. Li, M.~Choi, and F.~Zhang, ``Robot teleoperation for efficient pipe maintenance in nuclear facilities,'' in {\em Nuclear Plant Instrumentation and Control \& Human-Machine Interface Technology (NPIC\&HMIT 2025)}, 2025.

\bibitem{Do2024VRPlatform}
Y.~Do, C.-Y. Li, A.~Chokshi, and F.~Zhang, ``Enhanced virtual reality platform for training in nuclear power plants,'' in {\em International Symposium on Future Instrumentation and Control for Nuclear Power Plants}, 2024.

\bibitem{hesu2025preference}
A.~Hesu, S.~Kim, and F.~Zhang, ``Preference-based multi-robot planning for nuclear power plant online monitoring and diagnostics,'' {\em Nuclear Science and Engineering}, vol.~199, no.~8, pp.~1292--1309, 2025.

\bibitem{son2025physics}
H.~Son, Y.~Do, M.~Zebrowitz, and F.~Zhang, ``Physics-informed radiation multi-source localization with robotic platform: H. son et al.,'' {\em International Journal of Intelligent Robotics and Applications}, pp.~1--14, 2025.

\bibitem{fancyberbook}
F.~Zhang, ``Nuclear power plant cybersecurity,'' in {\em Nuclear power plant design and analysis codes}, pp.~495--513, Elsevier, 2021.

\bibitem{yoo2025self}
S.~Yoo, G.~Mohler, and F.~Zhang, ``Self-healing control of nuclear power plants under false data injection attacks,'' {\em Nuclear Science and Engineering}, vol.~199, no.~1, pp.~162--175, 2025.

\bibitem{chen2024full}
X.~Chen, J.~Coble, and F.~Zhang, ``A full-scope, high-fidelity simulator-based hardware-in-the-loop testbed for comprehensive nuclear power plant cybersecurity research,'' {\em Progress in Nuclear Energy}, vol.~175, p.~105348, 2024.

\bibitem{DBLP:journals/corr/BrockmanCPSSTZ16}
G.~Brockman, V.~Cheung, L.~Pettersson, J.~Schneider, J.~Schulman, J.~Tang, and W.~Zaremba, ``Openai gym,'' {\em CoRR}, vol.~abs/1606.01540, 2016.

\end{thebibliography}

\end{document}